\documentclass[journal]{IEEEtran}

\ifCLASSINFOpdf
\else
   \usepackage[dvips]{graphicx}
\fi
\usepackage{url}

\usepackage{grffile}
\usepackage{graphicx}
\usepackage{multirow}
\usepackage[sort,nocompress]{cite}
\usepackage{balance}
\usepackage{amsmath}

\begin{document}

\title{Improving Makeup Face Verification\\by Exploring Part-Based Representations}

\author{Marcus de Assis Angeloni and Helio Pedrini, \IEEEmembership{Senior Member, IEEE}
\thanks{The authors thank CAPES, FAPESP (grants \#2014/12236-1 and \#2017/12646-3), CNPq (grant \#309330/2018-1) for the financial support, and NVIDIA for the donation of a GPU as part of the GPU Grant Program.}
\thanks{M. A. Angeloni is with the Institute of Computing, University of Campinas (Unicamp), Campinas, SP, 13083-852, Brazil and with the AI R\&D Lab, Samsung R\&D Institute Brazil, Campinas, SP, 13097-160, Brazil (e-mail: marcus.angeloni@ic.unicamp.br).}
\thanks{H. Pedrini is with the Institute of Computing, University of Campinas (Unicamp), Campinas, SP, 13083-852 (e-mail: helio@ic.unicamp.br).}}

\maketitle

\begin{abstract}
Recently, we have seen an increase in the global facial recognition market size. Despite significant advances in face recognition technology with the adoption of convolutional neural networks, there are still open challenges, such as when there is makeup in the face. To address this challenge, we propose and evaluate the adoption of facial parts to fuse with current holistic representations. We propose two strategies of facial parts: one with four regions (left periocular, right periocular, nose and mouth) and another with three facial thirds (upper, middle and lower). Experimental results obtained in four public makeup face datasets and in a challenging cross-dataset protocol show that the fusion of deep features extracted of facial parts with holistic representation increases the accuracy of face verification systems and decreases the error rates, even without any retraining of the CNN models. Our proposed pipeline achieved 
competitive results for the four 
datasets (EMFD, FAM, M501 and YMU).
\end{abstract}


\IEEEpeerreviewmaketitle

\section{Introduction}

\IEEEPARstart{I}{n} recent years, we can observe an increase in the global facial recognition market size, mainly due to the growth of importance of the surveillance industry, investment by the government and defense sector and increasing technological advancement across industry verticals~\cite{MarketsAndMarkets}.
Despite significant advances in face recognition technology with the adoption of convolutional neural networks (CNN)~\cite{lightcnn_2018, vggface2_2018}, there are still open issues.
Among them, the use of facial makeup remains a challenge for automatic face recognition systems and even for human evaluations, since it is able to change the original face appearance and cover facial flaws~\cite{Wang2016, 10.5555/3298239.3298377, 8578110, ZHANG2019, emfd_2020}.

The preliminary work that explicitly established the impact of facial makeup on automated biometric systems was proposed a few years ago. 
Dantcheva et al.~\cite{ymu_2012, Chen2013} studied the impact of makeup in face verification task and showed that 
it can compromise the accuracy of a biometric system.
After that, Hu et al.~\cite{fam_2013} increased the accuracy in makeup scenario by measuring correlations between face images 
using canonical correlation analysis and applying support vector machines.

Since cosmetics are generally applied to facial components as eyes and mouth, an interesting direction of investigation 
is part-based representations. 
As demonstrated by the cognitive science literature, there is a high probability that part-based processing may exist in humans' perception of the face~\cite{Sagiv2001}.
Following this inspiration from our visual system, we can see a recent attention to 
this representation for face verification~\cite{Angeloni2016, 9211002}, facial gender and pose estimation~\cite{10.1145/3152125, Zavan2019}, facial age estimation~\cite{Yi2015, Angeloni2019} and facial expression~\cite{Jan2018, Happy2020}. 

Coupled with this growing attention to facial parts, a set of approaches were proposed to use this idea for makeup invariant face verification. 
Guo et al.~\cite{Guo2014} and Chen et al.~\cite{Chen2016} proposed two different patch scheme approaches, where 
local features are projected onto a subspace to make the match. 
Sun et al.~\cite{Sun2017} proposed a weakly supervised method pre-trained on Internet videos and fine-tuned on makeup datasets. In addition, they defined a new loss function 
and combined facial parts (and their mirroring) in the matching step using voting strategy. 
Li et al.~\cite{blan2019} proposed a bi-level adversarial network (BLAN), which used a generative adversarial network to generate non-makeup images from makeup ones preserving the person identity. To further improve the quality of their synthesized images, they used a two-path generator that consider global and local structure. 
More recently, Wang et al.~\cite{emfd_2020} proposed a unified multi-branch network which can simultaneously synthesize makeup faces through face morphology network (swapping local regions of different images) and learn cosmetics-robust face representations using attention-based multi-branch learning network. This multi-branch consisted of one holistic and three local branches (two eyes and mouth) that can capture complementary and detailed information.

\begin{figure*}[!htb]
	\begin{center}
		\includegraphics[width=.8\linewidth]{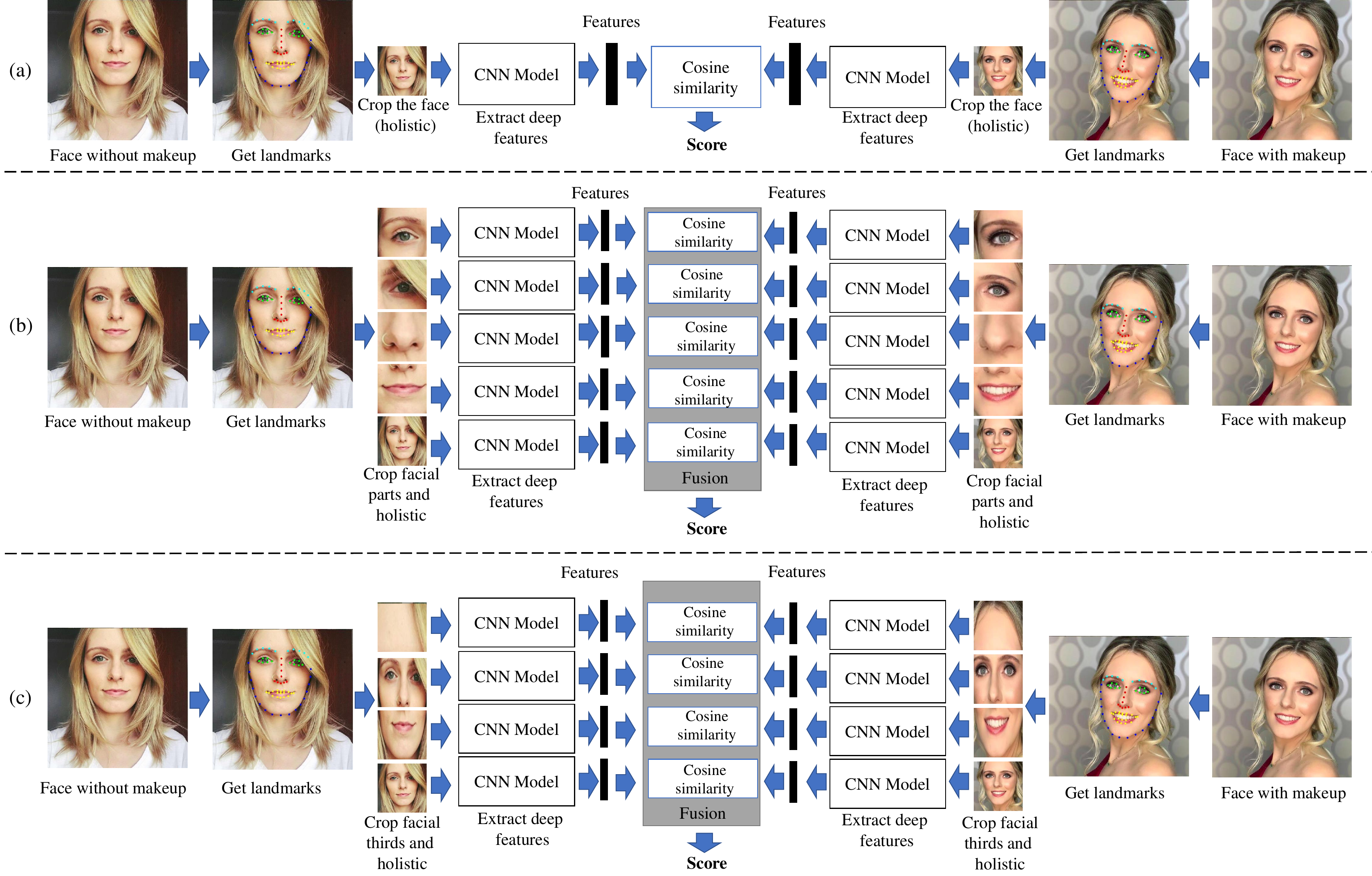}
	\end{center}
	\caption{Pipeline of the proposed part-based face verification: (a) holistic approach, used as baseline; (b) proposed part-based using 4 facial parts (left periocular, right periocular, nose and mouth) and holistic; (c) proposed part-based using 3 facial thirds (upper, middle and lower) and holistic.}
	\label{fig_pipeline}
\end{figure*}

In this letter, we propose a pipeline to improve the performance of face recognition with makeup by exploring part-based representations. 
First, we crop specific regions of the face analyzing two strategies: four parts located around fiducial points or splitting the face in facial thirds.  
Then, we extract features using publicly available state-of-the-art CNN models. 
Finally, we evaluate the fusion of holistic and facial region scores.
We show that significant improvements can be achieved by including facial parts, even without fine-tuning the CNN models.
The major contributions of this approach are: (i) proposition and evaluation of two strategies for cropping facial parts; (ii) performance comparison on face makeup datasets of CNN features extracted from the entire face (holistic) and fusing it with facial parts; (iii) evaluation of the generalization by exploring part-based representation in a cross-dataset protocol; (iv) complete reproducible experimental procedure, allowing the regeneration and extension of the obtained results.

The letter is organized as follows. 
Section~\ref{Method} presents the proposed method. 
Section~\ref{Experiments} reports the experimental settings and results. 
Finally, conclusions are drawn in Section~\ref{Conclusion}.

\section{Method}
\label{Method}
The main goal of this letter is to evaluate the adoption of facial parts in order to improve the accuracy in the challenging scenario of faces with makeup, by describing them with publicly available face models. 
The intuition behind the use of facial parts is that they suffer differently from the effects of cosmetics and together they can increase the accuracy and generalization of the existing face recognition systems. 
The overall pipeline of our method is illustrated in Fig.~\ref{fig_pipeline}.
In Fig.~\ref{fig_pipeline}a, the baseline of this letter is presented, where only the holistic face is used. 
In Fig.~\ref{fig_pipeline}b and Fig.~\ref{fig_pipeline}c, two strategies for cropping facial parts using facial landmarks coordinates and combining them with the holistic face are presented.

A face image without makeup and another with makeup are inputs to the pipeline.
The preprocessing step starts by applying a face detector followed by a 2D facial landmarks estimator, both available in DLib~\cite{dlib2009}. 
These landmarks are used to align, crop and resize the face thirds and facial parts.

The proposed facial parts 
differ from approaches that divide the face region into arbitrary blocks~\cite{7982752, YANG2020109}, as these blocks did not coincide or did not have their choice motivated by the fiducial regions.
Our first strategy for facial parts (Fig.~\ref{fig_pipeline}b) is composed of four components: left periocular, which includes the eye and eyebrow, right periocular, nose and mouth. 
These parts were chosen due to periocular region presented complementary information to face recognition~\cite{Tiago2015, LSP2020}, and nose and mouth regions were adopted in related works~\cite{Bonnen2013, Angeloni2016, Zavan2019, Angeloni2019, Jan2018}. 
We used the landmarks of each part and expand its borders until obtain a square region (expected as input to CNN models), applying a padding if it crosses the image boundaries. 
In this scheme, we maintain the aspect ratio and allow overlap between facial regions.
The second strategy for cropping facial parts (Fig.~\ref{fig_pipeline}c) is composed of three facial thirds, which were inspired by anthropometry studies~\cite{farkas1994anthropometry, 10.1145/280814.280823}, facial aesthetics~\cite{harrar2018art, EGGERSTEDT2020102643}
and facial beauty evaluation~\cite{milutinovic2014evaluation, LAURENTINI2014184}.
The upper third extends from the hairline to the glabella (point between eyebrows), the middle third from the glabella to the subnasale (base of the nose), and the lower third from the subnasale to the menton (end of chin).  Differently from the first approach, in this case the parts do not have any overlap between them and do not maintain the aspect ratio (due to the resize to square dimensions).

After the preprocessing, as illustrated in Fig.~\ref{fig_pipeline}, we directly extract features using the CNN models of each facial part/third and of the holistic face. 
To extract deep features, we chose 13 CNN models trained for face verification using four popular architectures. 
The first was the FaceNet~\cite{facenet_2015}, a CNN that provides a unified embedding, which maps each face image into an Euclidean space of 128 dimensions, such that the distances in that space correspond to face similarity. 
The second was the VGGFace~\cite{vggface_2015}, which was generated by training a VGG-16 CNN architecture~\cite{Simonyan15} using a very large scale face dataset. We extracted the deep features (4,096 dimensions) of the fully connected layer `fc7'. 
The third architecture was adopted in the VGGFace2 models~\cite{vggface2_2018}. 
VGGFace2 is a very large dataset with variations in pose, age, illumination and ethnicity used to train a set of CNN models using ResNet-50 architecture~\cite{Resnet2016}, with and without Squeeze-and-Excitation (SE) blocks~\cite{SENet2018}. 
In this letter, we evaluated eight models trained with VGGFace2. Four without SE blocks: ResNet-50, ResNet-50-128D (adding a 128-D embedding layer for feature representation), ResNet-50-256D (adding a 256-D embedding layer for feature representation), ResNet-50-FT (pre-trained on MS-Celeb-1M Dataset~\cite{msceleb2016}); and four with SE blocks: SE-ResNet-50, SE-ResNet-50-128D, SE-ResNet-50-256D, SE-ResNet-50-FT. 
The last evaluated architecture was the LightCNN~\cite{lightcnn_2018}. 
Its innovation was a variation of maxout activation, called max-feature-map that allows to obtain a 256-D face representation. In this letter, we explored three Light-CNN models, one called LightCNN-9, and the others based on LightCNN-29 architecture. It is important to mention that these models used grayscale face images instead of RGB images as inputs, according to their authors to alleviate the influence of illumination discrepancy. 
In our pipeline, we use cosine similarity among each pair as score, comparing the features extracted of the same facial part from different face images.

Finally, in order to fuse the scores computed by each facial part with the holistic score, 
we evaluate the sum rule strategy and two methods available in the Bob toolbox~\cite{bob2012,bob2017}: Linear Logistic Regression (LLR) and a Multi-Layer Perceptron (MLP) with three hidden layers.

\section{Experiments}
\label{Experiments}
In this section, we describe the datasets, evaluation protocols and experimental results shown as an ablation study.

\begin{figure}[!htb]
	\centering	
	{\includegraphics[width=0.14\linewidth]{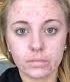}}
	{\includegraphics[width=0.14\linewidth]{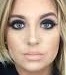}}
	\hspace{.1cm}
	{\includegraphics[width=0.14\linewidth]{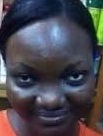}}
	\vspace{.1cm}	
	{\includegraphics[width=0.14\linewidth]{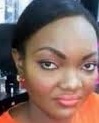}}
	\hspace{.1cm}
	{\includegraphics[width=0.14\linewidth]{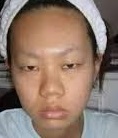}}
	{\includegraphics[width=0.14\linewidth]{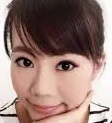}}
	\vspace{.1cm}
	{\includegraphics[width=0.14\linewidth]{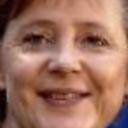}}
	{\includegraphics[width=0.14\linewidth]{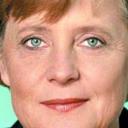}}
	\hspace{.1cm}
	{\includegraphics[width=0.14\linewidth]{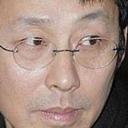}}	
	{\includegraphics[width=0.14\linewidth]{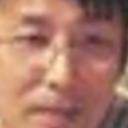}}
	\hspace{.1cm}
	{\includegraphics[width=0.14\linewidth]{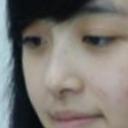}}
	{\includegraphics[width=0.14\linewidth]{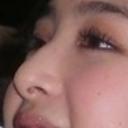}}
	\vspace{.1cm}
	{\includegraphics[width=0.14\linewidth]{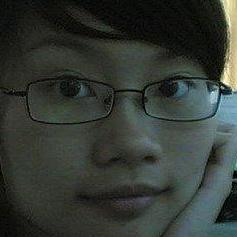}}
	{\includegraphics[width=0.14\linewidth]{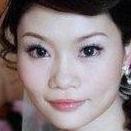}}
	\hspace{.1cm}
	{\includegraphics[width=0.14\linewidth]{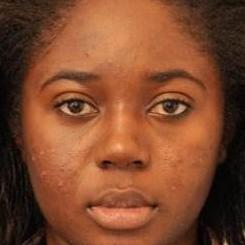}}	
	{\includegraphics[width=0.14\linewidth]{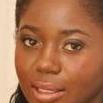}}
	\hspace{.1cm}
	{\includegraphics[width=0.14\linewidth]{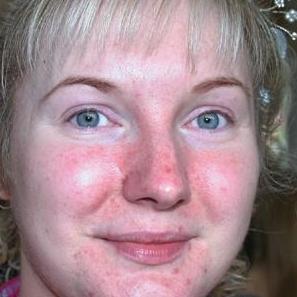}}
	{\includegraphics[width=0.14\linewidth]{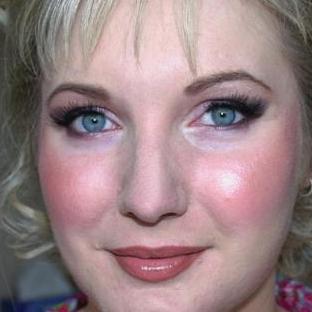}}
	\hspace{.1cm}
	{\includegraphics[width=0.14\linewidth]{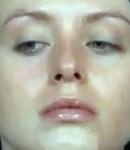}}
	{\includegraphics[width=0.14\linewidth]{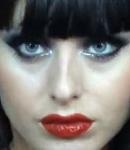}}
	\hspace{.1cm}
	{\includegraphics[width=0.14\linewidth]{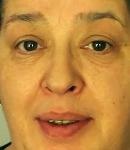}}	
	{\includegraphics[width=0.14\linewidth]{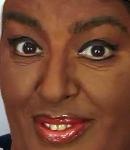}}
	\hspace{.1cm}
	{\includegraphics[width=0.14\linewidth]{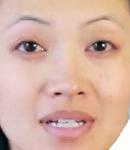}}
	{\includegraphics[width=0.14\linewidth]{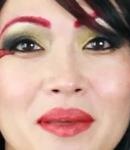}}
	
	\caption{Samples of makeup datasets with pairs of `no makeup' vs `makeup': EMFD (1$^{\text{st}}$ row), FAM (2$^{\text{nd}}$ row), M501 (3$^{\text{rd}}$ row) and YMU (4$^{\text{th}}$ row).}
	\label{fig:db}
\end{figure}

We used four datasets in our experiments. 
The Extended Makeup Face Dataset (EMFD)~\cite{emfd_2020} contains 551 pairs of face images. 
All images were collected from the Internet, where each pair has one makeup and one non-makeup face image of the same individual. 
This dataset includes males and females, some occlusions by glasses, variations in image dimensions, areas of acne and variations in head pose. 
The Face Makeup (FAM) Dataset~\cite{fam_2013} is composed of 519 pairs of face images available on the Internet, where 222 of them are males and 297 are females. 
Each pair is composed of one image with makeup and another without makeup. 
The M501 Dataset was collected in the work of Guo et al.~\cite{Guo2014}, which contains 501 pairs (one makeup face image and another without makeup) of female individuals collected from the Internet.  
The Youtube Makeup (YMU) Dataset~\cite{ymu_2012, Chen2016} is composed of 151 female subjects with images captured from YouTube makeup tutorials. 
Each subject has four images, two of them before the application of makeup and two after the application.
Some samples of these four datasets are shown in Fig.~\ref{fig:db}.

In order to verify whether adding features extracted from facial parts improves the face recognition system performance in scenarios with makeup, we adopted the following metrics~\cite{gunther2016face}: (i) Equal Error Rate (EER) is the point where false acceptance rate (FAR) and false rejection rate (FRR) intersect each other, i.e., where both rates are minimal and optimal in the evaluated system; (ii) Half Total Error Rate (HTER) is the average between FAR and FRR at a specific threshold; (iii) mean accuracy is an averaged result reported over $N$ rounds to measure the performance of algorithms.

\begin{table}
	\centering
	\caption{Results of holistic features vs. fusion with thirds and parts (in terms of EER (\%)). Best values marked in bold.}
	\small
	\setlength{\tabcolsep}{2.3pt}
	\begin{tabular}{l|rrrr}
		\hline
		\textbf{Features} & \textbf{EMFD} & \textbf{FAM} & \textbf{M501} & \textbf{YMU}  \\
		\hline
		FaceNet &  7.62 & 8.24 & 5.01 &  3.60 \\
		VGGFace & 14.70 & 20.42 & 13.62 & 8.13 \\
		VGGFace2 (ResNet-50-128D) & 5.26 & 7.71 & 3.74 & 2.94 \\
		VGGFace2 (ResNet-50-256D) & 5.60 &  7.55 & 3.99 & 2.96 \\
		VGGFace2 (ResNet-50-FT) & 5.26 & 8.08 & 2.99 & 2.65 \\
		VGGFace2 (ResNet-50) & 5.78 & 8.82 & 3.36 & 2.67 \\
		VGGFace2 (SE-ResNet-50-128D) & 4.90 & 7.71 & 3.60 & 3.19 \\
		VGGFace2 (SE-ResNet-50-256D) & 4.94 & 7.68 & 3.59 & 2.98 \\
		VGGFace2 (SE-ResNet-50-FT) & \underline{4.22} & 7.32 & 2.97 & 2.17 \\
		VGGFace2 (SE-ResNet-50) & 5.99 & 8.85 & 3.99 & 2.63 \\
		LightCNN-9 & 11.94 & 10.61 & 3.99 & 3.95 \\
		LightCNN-29 & 9.29 & 7.32 & 2.40 & 2.65 \\
		LightCNN-29v2 & 6.13 & \underline{5.78} & \underline{2.00} & \underline{1.48} \\
		\hline
		Third LLR fusion (same features)  & 4.34 & 5.58 & 1.80 & 1.32 \\
		Third LLR fusion (best combination) & 3.81 & 5.20 & \textbf{1.20} & \textbf{0.99} \\
		Third MLP fusion (same features)  & 14.70 & 14.22 & 12.94 & 9.44 \\
		Third MLP fusion (best combination) & 4.55 & 5.42 & 2.02 & 1.82 \\
		Third Sum fusion (same features)  & 13.21 & 12.74 & 6.54 & 5.46\\
		Third Sum fusion (best combination) & 10.09 & 11.95 & 4.39 & 4.11 \\		
		\hline
		Part LLR fusion (same features) & 3.99 & 5.37 & 1.60 & 1.49 \\
		Part LLR fusion (best combination) & \textbf{3.63} & \textbf{4.83} & \textbf{1.20} & \textbf{0.99} \\
		\hline

	\end{tabular}
	\label{tab_RESULTS}
\end{table}

We started our experiments by evaluating the performance of the features in the datasets separately, in terms of EER. Table~\ref{tab_RESULTS} presents the comparative results between the baseline results, i.e., holistic features (13 first rows) and the fusion results with facial thirds and parts (last 8 rows). 
LightCNN-29v2 features achieved the lowest EER in the experiments using only holistic face for FAM, M501 and YMU datasets. The exception was EMFD, in which VGGFace2 (SE-ResNet-50-FT) features were the best. 
The EER of facial crops individually was higher than 15\% for facial thirds and higher than 25\% for facial parts. 
Although these results of facial crops are poor, when we fuse them with holistic we achieved significant improvements. 
We evaluate three strategies to fuse the facial third scores and the holistic representation: Sum, LLR and MLP fusion. 
Our fusion experiments use only one system for each facial third in the combination, that is, we combine each one of the 13 deep features of upper third with each one of the middle, lower third and holistic, adopting three ways of fusion. 
Since the LLR achieves the best results in the fusion experiments with facial thirds, we use only this strategy in the fusion experiments with facial parts. 

We reported two results for each fusion strategy, the best result adopting the same features for all regions (showing improvement even when we consider a biometric system as a black box), and the best result allowing that each region use a different feature.
A significant improvement was achieved for all datasets, as can be seen in the rows with ``best combination'' reported in Table~\ref{tab_RESULTS}. 
In the fusion results, facial parts were better or equal than facial thirds.

To check generalization, we adopted a new protocol where the threshold is defined on a dataset (EER point) and the performance is computed on all datasets using the same threshold in terms of HTER (cross dataset). Table~\ref{tab_cross} shows the results of this cross-dataset protocol. The datasets are shown in the rows, where the threshold and weights of the LLR fusion are defined. The HTER of the four datasets are shown from the second to fourth column, and their average, standard deviation and maximum (worst result in the generalization) in the last three columns. 
The features used in the experiments were the same reported as best results in Table~\ref{tab_RESULTS}. 
Again, reductions in error rates are observed when fusing with the facial parts and thirds. 
However, this time the best results were achieved with the fusion of holistic with facial thirds. 

\begin{table}
	\centering
	\caption{Cross-dataset protocol (in terms of HTER (\%)).}
	\small
	\renewcommand{\arraystretch}{0.99}
	\setlength{\tabcolsep}{3pt}
	\begin{tabular}{l|rrrr|rr}
		\hline
		\textbf{Holistic} & \textbf{EMFD} & \textbf{FAM} & \textbf{M501} & \textbf{YMU} & \textbf{Avg. $\pm$ S.D.} & \textbf{Max.} \\
		\hline
		\textbf{EMFD} & 4.22 & 7.40 & 3.90 & 3.04 & 4.64 $\pm$ 1.91 & 7.40 \\
		\textbf{FAM} & 6.18 & 5.78 & 2.17 & 2.40 & 4.13 $\pm$ 2.14 & 6.18 \\
		\textbf{M501} & 6.87 & 5.46 & 2.00 & 1.98 & 4.08 $\pm$ 2.48 & 6.87 \\
		\textbf{YMU} & 7.68 & 5.25 & 1.76 & 1.48 & 4.04 $\pm$ 2.97 & 7.68 \\
		\hline
		\hline	
		\textbf{Part fusion} & \multirow{2}{*}{\textbf{EMFD}}  & \multirow{2}{*}{\textbf{FAM}} & \multirow{2}{*}{\textbf{M501}} & \multirow{2}{*}{\textbf{YMU}} & \multirow{2}{*}{\textbf{Avg. $\pm$ S.D.}} & \multirow{2}{*}{\textbf{Max.}} \\
		\textbf{(same feat.)} &  &  &  &  &  &  \\
		\hline
		\textbf{EMFD} & 3.99 & 7.40 & 2.90 & 2.16 & 4.11 $\pm$ 2.32 & 7.40 \\
		\textbf{FAM} & 6.18 & 5.37 & 3.05 & 2.45 & 4.26 $\pm$ 1.79 & 6.18 \\
		\textbf{M501} & 6.75 & 5.95 & 1.60 & 1.78 & 4.02 $\pm$ 2.71 & 6.75 \\
		\textbf{YMU} & 6.05 & 5.56 & 2.10 & 1.49 & 3.80 $\pm$ 2.34 & 6.11 \\	
		\hline
		\textbf{Part fusion} & \multirow{2}{*}{\textbf{EMFD}}  & \multirow{2}{*}{\textbf{FAM}} & \multirow{2}{*}{\textbf{M501}} & \multirow{2}{*}{\textbf{YMU}} & \multirow{2}{*}{\textbf{Avg. $\pm$ S.D.}} & \multirow{2}{*}{\textbf{Max.}} \\
		\textbf{(best comb.)} &  &  &  &  &  &  \\		
		\hline
		\textbf{EMFD} & \textbf{3.63} & 6.68 & 2.96 & 2.33 &  3.90 $\pm$ 2.23 & 6.68 \\
		\textbf{FAM} & 6.11 & \textbf{4.83} & 2.82 & 2.30 & 4.02 $\pm$ 1.94 & 6.11 \\
		\textbf{M501} & 7.44 & 6.60 & \textbf{1.20} & 1.52 &  4.19 $\pm$ 3.03 & 7.44 \\
		\textbf{YMU} & 6.44 & 5.92 & 1.78 & \textbf{0.99} &  3.78 $\pm$ 3.01 & 6.44 \\
		\hline
		\hline
		\textbf{Third fusion} & \multirow{2}{*}{\textbf{EMFD}}  & \multirow{2}{*}{\textbf{FAM}} & \multirow{2}{*}{\textbf{M501}} & \multirow{2}{*}{\textbf{YMU}} & \multirow{2}{*}{\textbf{Avg. $\pm$ S.D.}} & \multirow{2}{*}{\textbf{Max.}} \\
		\textbf{(same feat.)} &  &  &  &  &  &  \\
		\hline
		\textbf{EMFD} & 4.34 & 6.79 & 3.06 & 2.31 & 4.13 $\pm$ 1.96 & 6.79 \\
		\textbf{FAM} & 5.72 & 5.58 & 3.13 & 2.13 & 4.14 $\pm$ 1.79 & 5.72 \\
		\textbf{M501} & 6.32 & 6.41 & 1.80 & 1.43 & 3.99 $\pm$ 2.75 & 6.41 \\
		\textbf{YMU} & 5.91 & 5.56 & 1.89 & 1.32 &  3.67 $\pm$ 2.40 & 5.91 \\	
		\hline
		\textbf{Third fusion} & \multirow{2}{*}{\textbf{EMFD}}  & \multirow{2}{*}{\textbf{FAM}} & \multirow{2}{*}{\textbf{M501}} & \multirow{2}{*}{\textbf{YMU}} & \multirow{2}{*}{\textbf{Avg. $\pm$ S.D.}} & \multirow{2}{*}{\textbf{Max.}} \\
		\textbf{(best comb.)} &  &  &  &  &  &  \\
		\hline
		\textbf{EMFD} & 3.81 & 6.54 & 2.66 & 2.26 & 3.82 $\pm$ 1.93 & 6.54 \\
		\textbf{FAM} & 5.64 & 5.20 & 3.10 & 2.09 & 4.01 $\pm$ 1.69 & \textbf{5.64} \\
		\textbf{M501} & 6.96 & 5.82 & \textbf{1.20} & 1.37 & 3.84 $\pm$ 2.98 & 6.96 \\
		\textbf{YMU} & 6.17 & 5.45 & 1.62 & \textbf{0.99} &  \textbf{3.56} $\pm$ 2.63 & 6.17 \\
		\hline
	\end{tabular}
	\label{tab_cross}
	\vspace{-.34cm}
\end{table}

After verifying the gains obtained by combining the scores of holistic approaches with that based on parts, we evaluated the performance of the best combinations with results available in the literature. 
To be fair, we adopted the same protocol and metrics defined in the original publications of the datasets. 
The protocol of YMU dataset~\cite{ymu_2012,Chen2016} evaluates the performance of non-makeup vs makeup in terms of EER. Our results presented in Table~\ref{tab_RESULTS} showed error rates significantly smaller than previous works~\cite{Chen2013_LGGP, Chen2016} following their protocol. We did not compare with more recent works~\cite{SAEED2021106921, roy2021local}, since they had proposed their own protocols for YMU.
The protocol for other three datasets (EMFD, FAM and M501) considers only non-makeup vs makeup reported in terms of mean accuracy of 5 folds, using the same number of positive and negative trials in the test set.
The comparison of the proposed results with literature is presented in Table~\ref{tab_other}. 
Our pipeline achieved competitive results in all datasets (with improvements when fusing with facial parts) even without any retraining of the CNN models with makeup face trials as the competitors.

\begin{table}
	\centering
	\caption{Results following the protocol of the EMFD, FAM and M501 datasets, in terms of mean accuracy (\%) of 5-folds. Our best results marked in underline.}
	\small
	\setlength{\tabcolsep}{3pt}
	\begin{tabular}{l|ccc}
		\hline
		\textbf{Method} & \textbf{EMFD} & \textbf{FAM} & \textbf{M501}  \\
		\hline
		Hu et al.~\cite{fam_2013} & - & 59.60 & - \\
		Guo et al.~\cite{Guo2014} & - & - & 80.50 \\
		Sun et al.~\cite{Sun2017} & - & - & 82.40 \\
		Li et al.~\cite{blan2019} & - & 88.10 & 94.80 \\
		Wang et al.~\cite{emfd_2020} & \textbf{96.56} & \textbf{90.43} & \textbf{98.12} \\
		\hline 
		Best holistic & 91.10 & 87.47 & 96.21 \\
		Part fusion (same features) & 91.29 & 89.59 & 96.41 \\
		Part fusion (best combination) & \underline{92.56} & \underline{89.79} & \underline{97.21} \\
		Third fusion (same features)  & 90.20 & 87.86 & 96.21 \\
		Third fusion (best combination) & \underline{92.56} & 89.40 & 96.41 \\
		\hline
	\end{tabular}
	\label{tab_other}
	\vspace{-.34cm}
\end{table}

\begin{figure}[!htb]
	\centering	
	{\includegraphics[width=0.14\linewidth]{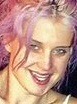}}
	{\includegraphics[width=0.14\linewidth]{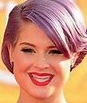}}
	\vspace{.1cm}
	\hspace{.1cm}
	{\includegraphics[width=0.14\linewidth]{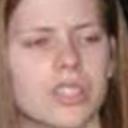}}
	{\includegraphics[width=0.14\linewidth]{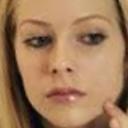}}
	\hspace{.1cm}
	{\includegraphics[width=0.14\linewidth]{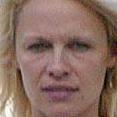}}
	{\includegraphics[width=0.14\linewidth]{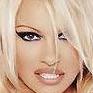}}
	
	{\includegraphics[width=0.14\linewidth]{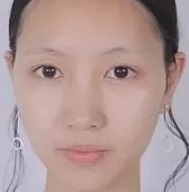}}
	{\includegraphics[width=0.14\linewidth]{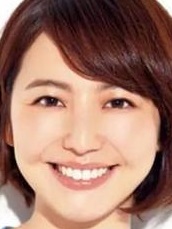}}
	\hspace{.1cm}
	{\includegraphics[width=0.14\linewidth]{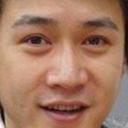}}
	{\includegraphics[width=0.14\linewidth]{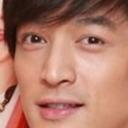}} 
	\hspace{.1cm}
	{\includegraphics[width=0.14\linewidth]{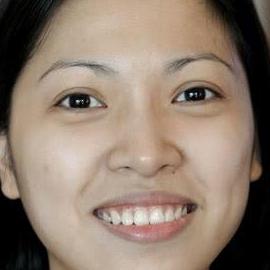}}
	{\includegraphics[width=0.14\linewidth]{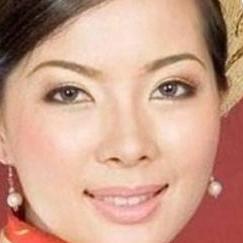}}
	\caption{Examples of misclassification of the holistic approach solved by fusing with facial parts: false rejection (first row) and false acceptance (second row).}
	\label{fig:ablationmakeup}
\end{figure}

In Fig.~\ref{fig:ablationmakeup} we present some examples of misclassifications in the best holistic system for each makeup dataset that are fixed when we fuse them with facial parts, showing the potential of our proposed approach.

\vspace{-.32cm}

\section{Conclusion}
\label{Conclusion}
In this letter, we proposed two strategies for cropping facial parts and evaluate the improvements of these adoption for the face verification with makeup task.
Experiments performed in four publicly available datasets and also in a cross-dataset protocol showed that improvements can be achieved in the evaluation metrics, even without any retraining or fine-tuning of the CNN models. 
Since we provided the source code of the proposed pipeline\footnote{The source data will be available online upon this letter acceptance.}, the entire experimental procedure can be reproduced to regenerate and extend the obtained results. 
Some viable future research topics include the investigation of other features to represent facial parts, and the evaluation of facial parts for other challenges in face verification, such face images with occlusion and noise.

\vfill\pagebreak
\balance
\bibliographystyle{ieee}
\bibliography{references}

\end{document}